\documentclass{bmvc2k}
\usepackage{times}
\usepackage{epsfig}
\usepackage{graphicx}
\usepackage{amsmath}
\usepackage{amssymb}
\usepackage{rotating}
\usepackage{caption}
\hypersetup{colorlinks}
\definecolor{bmvcblue}{RGB}{0, 0, 0}
\usepackage{calrsfs}

\title{Deep Interactive Region Segmentation and Captioning}

\addauthor{\small{Ali Sharifi Boroujerdi, Maryam Khanian \& Michael Breu\ss}}{}{1}

\addinstitution{\small{
Brandenburg University of Technology, Cottbus-Senftenberg, Germany
}}

\runninghead{ }{}

\begin{document}

\maketitle

\begin{abstract}
With recent innovations in dense image captioning, it is now possible to describe every object of the scene with a caption while objects are determined by bounding boxes. However, interpretation of such an output is not trivial due to the existence of many overlapping bounding boxes. Furthermore, in current captioning frameworks, the user is not able to involve personal preferences to exclude out of interest areas. In this paper, we propose a novel hybrid deep learning architecture for interactive region segmentation and captioning where the user is able to specify an arbitrary region of the image that should be processed. To this end, a dedicated Fully Convolutional Network (FCN) named Lyncean FCN (LFCN) is trained using our special training data to isolate the User Intention Region (UIR) as the output of an efficient segmentation. In parallel, a dense image captioning model is utilized to provide a wide variety of captions for that region. Then, the UIR will be explained with the caption of the best match bounding box. To the best of our knowledge, this is the first work that provides such a comprehensive output. Our experiments show the superiority of the proposed approach over state-of-the-art interactive segmentation methods on several well-known datasets. In addition, replacement of the bounding boxes with the result of the interactive segmentation leads to a better understanding of the dense image captioning output as well as accuracy enhancement for the object detection in terms of Intersection over Union (IoU).
\end{abstract}

\section{Introduction}
\label{sec:intro}
\begin{figure}
\begin{tabular}{ccccc}
\hspace{-2mm}
\includegraphics[width=2.2cm]{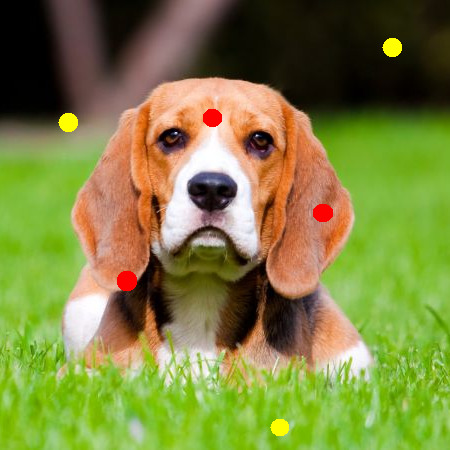}
&
\hspace{-3mm}
\includegraphics[width=2.2cm]{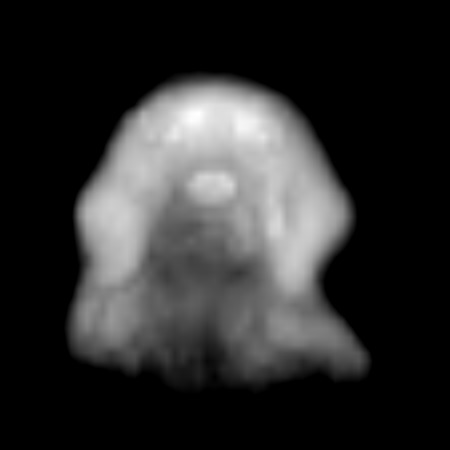}
&
\hspace{-3mm}
\includegraphics[width=2.2cm]{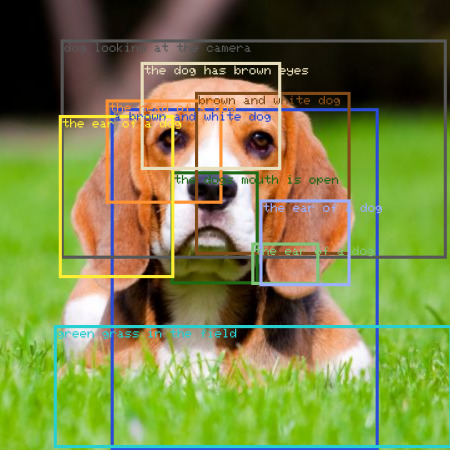}
&
\hspace{-3mm}
\includegraphics[width=2.2cm]{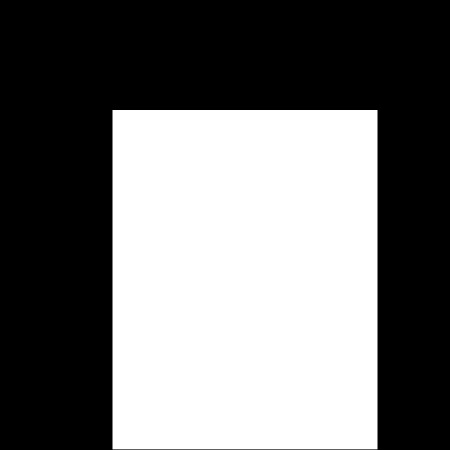}
&
\hspace{-3mm}
\includegraphics[width=2.2cm]{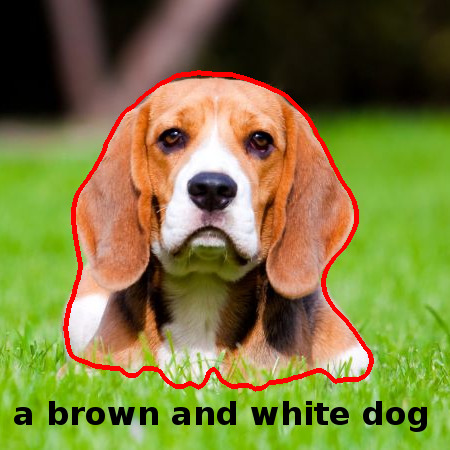}
\\
(a)&(b)&(c)&(d)&(e)
\end{tabular}
\vspace{2mm}
\caption{\small{Input image including positive and negative user clicks (a), probability map of our LFCN considering user intention (b), output of the dense image captioning \cite{dense} (c), best match bounding box w.r.t. user intention (d), and the final output of the proposed model including determined user intention region (UIR) and its description (e).}}
\label{fig1}
\vspace{-5mm}
\end{figure}

As one of the main sources of the human knowledge, our visual system including eyes, optic nerves and brain is able to easily detect, separate and describe each object of a scene. Inspired by this natural ability, interactive region segmentation and captioning is the task of parallel detection, separation and description of the visual user interests. This procedure can be exploited in several complex applications such as automatic image annotation and retrieval \cite{tag1,tag2}. To approach the task, one needs to have a full understanding of the scene which is equivalent to recognize and also locate all the visible objects. To this end, several object recognition techniques \cite{D40,D14,D46} have been proposed to detect image objects in different scales. In most of the literature, detected objects are determined by drawing bounding boxes around them. Although this notation is able to facilitate the detection process by decreasing its computational complexity, such an output is less informative when dealing with geometrical properties of the objects. As a more illustrative visual recognition technique, semantic segmentation \cite{I17,I3,I16,IS1} aims to assign a label to each pixel of the image where the labels can be class-aware or instance-aware. While the multi-level nature of semantic segmentation increases the problem dimensionality, interactive image segmentation \cite{GrabCut,I1,I8} tries to adjust the segmentation task with the user priorities in a simpler problem space. In reality, it sounds reasonable that human users may have a more restricted area of interest than the entire scope of the scene. Thus, the multi-dimensional semantic segmentation task can be shrunk to a binary segmentation problem aiming to separate the User Intention Region (UIR) as the foreground from other parts of the scene which requires less time and computations. 
\\
Equipped by the rich semantic memory of the visual data, the human observer is easily able to provide detailed explanation about different parts of an image which is a hard task in artificial intelligence. Thanks to recent developments of language models \cite{LM10, D21}, image captioning \cite{D32,D49,D24,D51,D8,D4,D11} makes it possible to produce linguistic descriptions of an image through a multimodal embedding of the visual stimuli and the word representation \cite{WE13} in a joint vector space \cite{D21}.
\\
In this paper, we propose a novel hybrid deep architecture for integrated detection, segmentation and captioning of the user preferences where the amount of the user interactions is limited to one or a few clicks. To this end, we designed a heuristic technique for the efficient generation of the synthetic user interactions. In addition, the new architecture of the proposed Lyncean Fully Convolutional Network (LFCN) leads to a better sight of the deep component that is responsible for interactive segmentation. Last but not least, as depicted in Fig. \ref{fig1}, our combination of interactive segmentation and dense captioning tasks introduces a new class of outputs where the user intention recognition meets linguistic interpretations and vice versa. Let us stress at this point that our main contributions are (i) to provide the first deep framework for combined interactive segmentation and captioning, and (ii) to achieve segnificant improvements in the interactive segmentation over other methods.
 
\section{More Details on Related Works}
With the increasing popularity of deep learning architectures \cite{D26,D53,D45}, both detection and captioning procedures have attracted a new wave of considerations. Convolutional Neural Networks (CNNs) \cite{D29,D26} have presented the ability to construct numerous visual features in different levels of abstraction through supervised learning. This property leads to feature generators that are able to reach near-human performance in various computer vision tasks \cite{resnet}. In addition, the structure of Fully Convolutional Networks (FCNs) \cite{I17} made it feasible to apply inputs of any size to the network and generate associated output in the same spatial domain. In contrast to CNNs, FCNs are able to maintain spatial information which is crucial to perform a pixel-level prediction such as semantic segmentation, object localization \cite{D40}, depth estimation \cite{depth} and interactive segmentation. Furthermore, Recurrent Neural Networks (RNNs) \cite{D50,D18} reveal potential for learning long term dependencies which is essential for simulating the continuous space of natural languages.
\\
Recently, CNN-RNN models are proposed to wrap detection and captioning tasks in an end-to-end learnable platform \cite{dense}. However, up to now the results appear to be mostly an unorganized and overcrowded set of captions and bounding boxes. These results are not easily understandable especially in the presence of several overlapping region proposals cf. Fig. \ref{fig1} (c). In addition, they do not involve user intentions.
\\
Before the success of CNNs in object detection, some classical techniques such as Histogram of Gradients (HoG) \cite{HoG}, Deformable Part Models (DPM) \cite{DPM} and selective search \cite{ss} (as an explicit region proposal method) were proposed. Later, in the Region-based CNN (R-CNN) model \cite{D14}, each proposed region has been forwarded through a separate CNN for the feature extraction. This model had some drawbacks such as a complex multi-stage training pipeline and expensive training process. To overcome those obstacles, the Fast R-CNN model \cite{D13} is proposed where a combination of a CNN and the Region of Interest (RoI) pooling mechanism is used to produce better information for region proposal. In the Faster R-CNN \cite{D38}, the CNN architecture is used not only for the feature extraction but also for region proposal itself. This leads to the invention of the Region Proposal Networks (RPNs) that are able to share full-image convolutional features with detection networks. The main achievement of this innovation is the parallel detection and localization of the objects in one forward pass of the network. In spite of these improvements, such models are not able to backpropagate through the bounding boxes information. Recently, Johnson et. al \cite {dense} proposed a localization layer based on Faster R-CNN architecture where the RoI pooling mechanism is replaced by bilinear interpolation \cite{D16,D19} that makes it possible to propagate backward through all the information.
\\ 
The primary purpose of the image captioning was image annotation \cite{IC1,K48} as the automatic assignment of some keywords to a digital image. By replacing keywords with some sentences that are able to describe not only the image objects but also the semantic relations in between, image captioning received more attention. The main problem in the automatic image description was the scarcity of the training data. Recent development of large datasets including images and their descriptions \cite{I13,IC2,D25} makes it feasible to expand learning-based captioning techniques. Classical image captioning approaches produced image descriptions by generative grammars \cite{IC3,IC4} or pre-defined templates working on some specific visual features \cite{IC5,IC6}.  In contrast, recently developed deep learning solutions apply an RNN-based language model that is conditioned on the output vectors of a CNN to generate descriptive sentences \cite{IC7,IC8,D49,D4,D21,D8}.
\\
With the growing popularity of interactive devices such as smart phones and tablets, interactive image processing attracts more attention. Interactive segmentation offers a pixel-wise classification based on user priorities. Among all the traditional approaches of the interactive segmentation, stroke-based techniques \cite{I22,I11,I26} are often based on graph cut techniques. In these methods, an energy function based on region/boundary division is optimized to find the segmentation result. Alternative approaches include random walks \cite{I8} and geodesics \cite{I1,I4}, mostly relying on low-level features such as color, texture and shape information. These types of attributes can be difficult to apply when the image appearance is complicated due to complex lighting conditions or existence of intricate textures. Recently, deep learning models have been used for interactive segmentation where the information of the image will be considered in higher semantic levels. To this end, FCNs as the standard frameworks for pixel-wise end-to-end learning tasks, have been applied \cite{I29, IS1,IS2,iter}.

\section{Proposed Method}  
\begin{figure*}
\begin{center}
\begin{tabular}{ccc}
\hspace{-2mm}\includegraphics[width=0.32\linewidth]{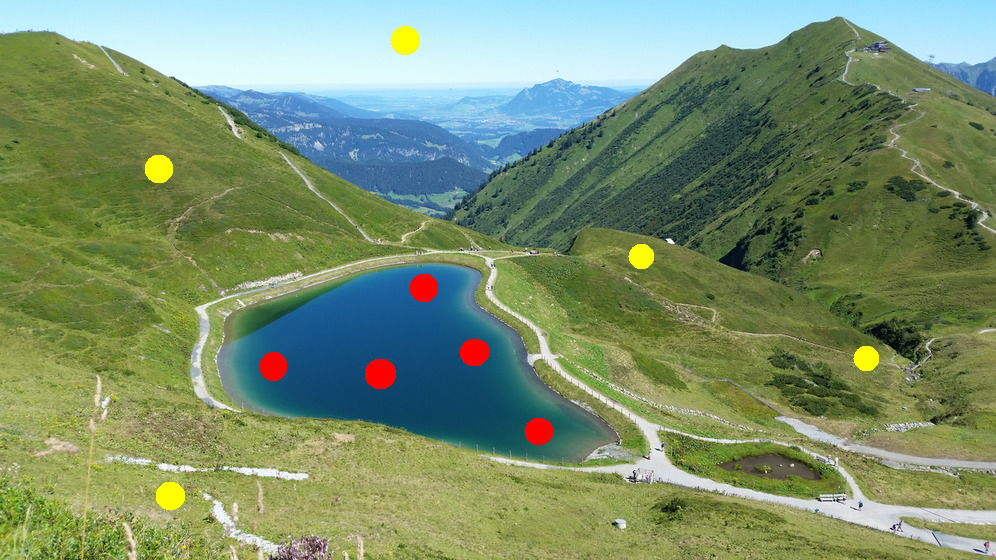}
&
\hspace{-2mm}\includegraphics[width=0.32\linewidth]{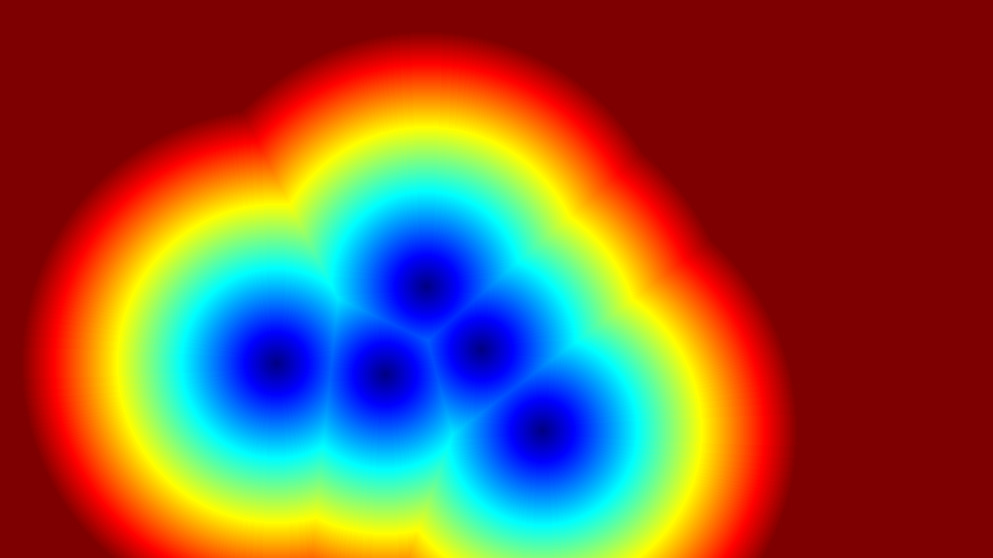}
&
\hspace{-2mm}\includegraphics[width=0.32\linewidth]{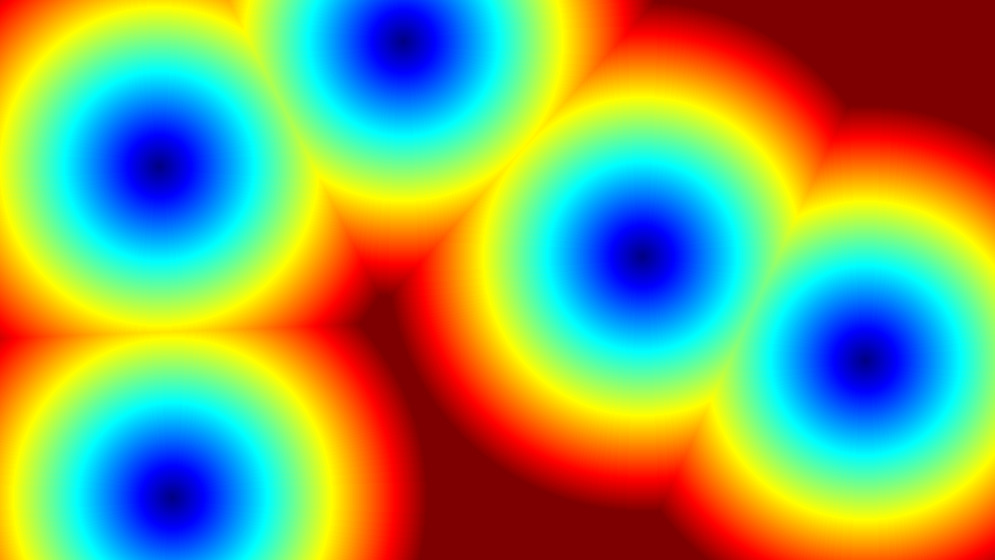}
\end{tabular}
\end{center}
   \caption{\small{Input image including positive and negative clicks (left), and obtained positive (middle) and negative (right) Voronoi diagrams.}}
\label{fig2}
\vspace{-5mm}
\end{figure*}

Our model receives an input image as well as user interactions in the form of positive/negative clicks and provides a seamless framework to generate accurate segmentation as well as expressive description of the UIR. In the preprocessing step, an efficient morphological technique will be used to provide  a huge amount of training samples in the form of synthetic user interactions. Then, each set of positive/negative seeds will be transformed into separate Voronoi diagrams as shown in Fig. \ref{fig2}. Next, a sequence of dedicated LFCNs with different granularities are applied as the interactive segmentation modules. Afterwards, a dense captioning architecture inspired by \cite{dense} will be utilized to obtain a number of region proposals along with their captions. In the fusion step, a heuristic method will be provided to combine results of localization, segmentation and captioning procedures to acquire highlighted borders of the UIR along with its expositor caption. In the following, we will investigate all steps of our model in detail.
\subsection{User Action Imitation}
\begin{figure}[!h]
\begin{center}
\vspace{3mm}
\begin{tabular}{c}
\includegraphics[width=0.65\linewidth]{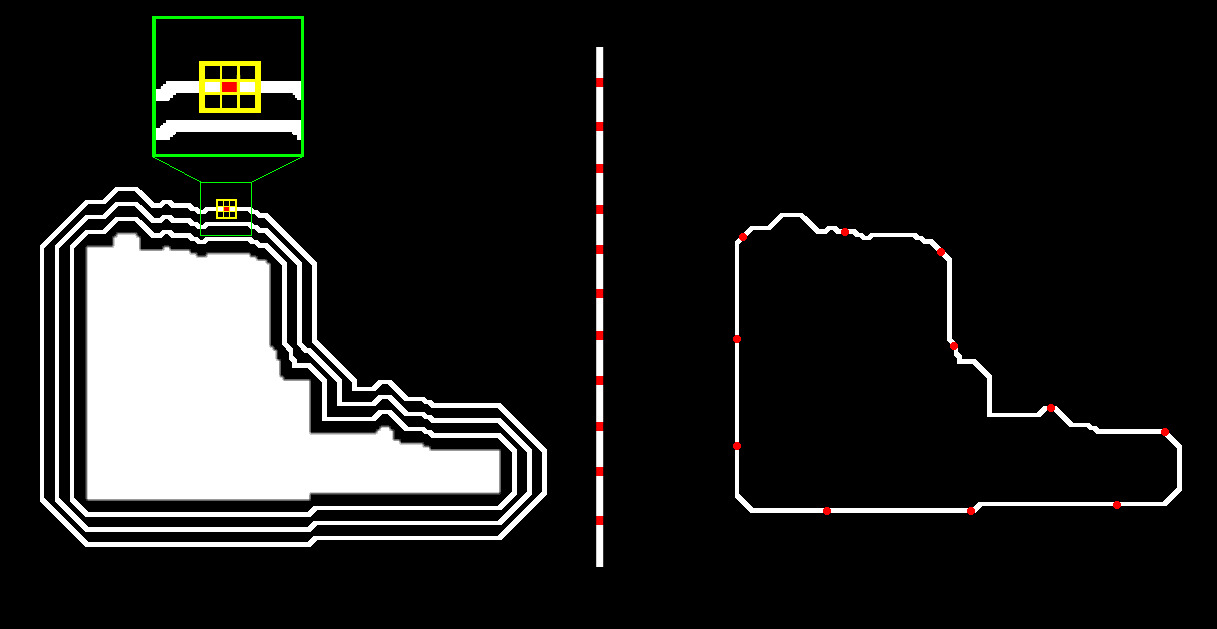}
\end{tabular}
\end{center}
   \caption{Our proposed Morphological Cortex Detection (MCD) technique.}
\label{mcd}
\vspace{-1mm}
\end{figure}

During interactive segmentation, the user will be asked to provide some general information about the position of the intended region. The requested information consists of some positive and negative seeds as depicted in Fig. \ref{fig2} which are equivalent to internal and external points of the UIR, respectively. Next, each set of seeds will be used to shape a Voronoi diagram. We denote each seed by $s_k$, $k=\{1,\ldots,n\}$. The value of pixel $v_{i,j}$ of the Voronoi diagram will be calculated by
\begin{equation}
v_{i,j}:=\min \{D_1,D_2,\dots,D_{n}\}
\end{equation}
where $D_k$ is the Euclidean distance of $v_{i,j}$ to the seed $s_k$.
To summarize, the value of each pixel in the Voronoi diagram is the Euclidean distance of that pixel to the nearest seed. For the sake of clarity, there should be a minimum inter-cluster distance in each set of the seeds. In addition, a minimum intra-cluster distance is also required to retain boundary regions of the clusters as distinctive as possible. So:
\\
\begin{itemize}
\item Every pair of seeds in each set should preserve a pre-defined distance from each other:
\begin{equation}
\exists \,d_1\in  \mathbb{R^+} : \forall (s_i,s_j)\in S, \|(s_i,s_j)\|_2>d_1
\end{equation}
\item All the seeds of each set should preserve a minimum distance from boundary pixels of the UIR ($\partial(UIR)$):
\begin{equation}
\exists \,d_2 \in \mathbb{R^+} :\forall s_i\in S, u\in \partial(UIR), \|(s_i,u)\|_2>d_2
\end{equation}
\end{itemize}
\noindent As expected, natural collection of such a data is unreasonably time consuming and expensive. Recently, Xu et al. \cite{iter} proposed some strategies for synthetic generation of user interactions. They ordained random generation for positive clicks inside the UIR while three distinct set of negative clicks are chosen as: 1) random background pixels with a certain distance to the UIR, 2) a point cloud inside the negative objects and 3) a uniform set of surrounding points of the UIR. Since their implementation is not publicly available, it seems their first and the second negative strategies do not obey natural interactions and the third one may be computationally expensive (see equation (2) in \cite{iter}). 
\\
\noindent
\textbf{Morphological Cortex Detection (MCD).} While the inside of the UIR can be quite small, the background region is usually large enough to provide useful geometric information about the UIR. Consequently, it is beneficial to generate negative seeds that surround the UIR uniformly. To provide an efficient implementation for such an interaction, we replace third negative strategy proposed in Xu et al. \cite{iter} with a Morphological Cortex Detection (MCD) technique that noticeably improves computational efficiency. Moreover, this method is able to simulate UIR cortex in different scales that enables convolutional filter of the LFCN to track UIR geometry in different layers. To implement this idea, a 1-pixel-wide boundary shape of the UIR will be extracted by performing a dilation on the binary mask of UIR in training dataset. Then, the original mask is subtracted from the dilation result. In the next step, this boundary path will be completely traversed using a $3\times3$ window to transfer all the boundary points' coordinates into a 1-D array in which the requested negative seeds can be selected uniformly. As the result of the MCD process, a uniform set of negative seeds will be obtained that represents the cortex of the UIR perfectly. The visual illustration of this technique is shown in Fig. \ref{mcd}. During our experiments, positive clicks are simulated randomly inside the UIR while negative seeds are generated by MCD mechanism in three different levels. 
\subsection{Intention Recognition}
For the task of intention recognition, we make use of a dedicated version of the standard FCN \cite{I17} where the last two fully connected layers are replaced with three convolutional layers containing decreasing kernel sizes of 7, 5 and 3. The impact of such an alternation is the gradual growth of the receptive field. This property improves network recognition of objects' geometry. Hence, we named this architecture as Lyncean Fully Convolutional Network (LFCN). By a proper use of zero padding, all the extended convolutional layers have the same output size. At the end of the extended part, the aggregated output of the additional layers will be upsampled to the size of the input as elaborated in \cite{I17}. 
\subsection{Fusion Approach}
\begin{figure*}
\begin{center}
\begin{tabular}{c}
\hspace{-2mm}\includegraphics[width=1\linewidth]{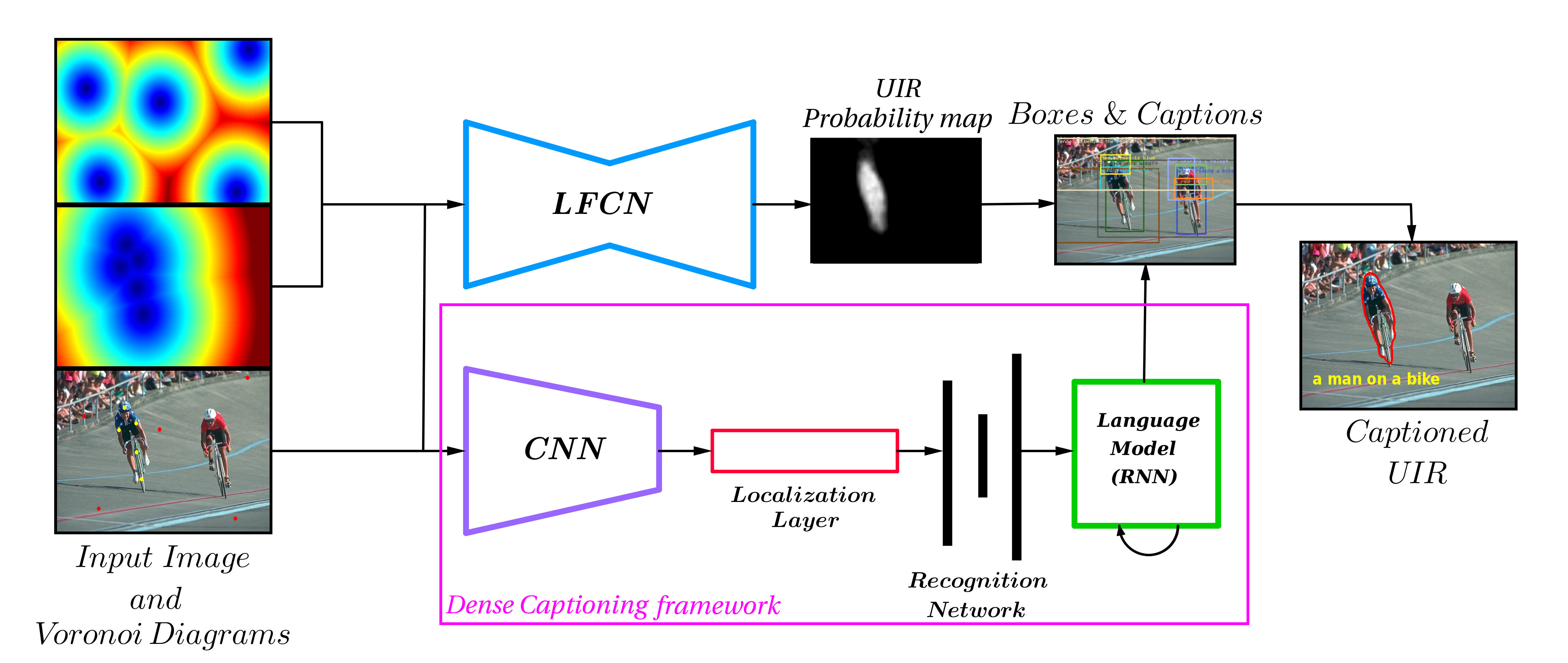}
\end{tabular}
\end{center}
   \caption{\small{Architecture of the proposed deep interactive region segmentation and captioning model.}}
\label{fig3}
\vspace{-5mm}
\end{figure*}

In order to supplement the result of the interactive segmentation with a proper linguistic commentary, we employ the dense image captioning framework \cite{dense}. The internal RPN of this architecture provides confidence scores for the existence of the object in proposed regions. After descending sort of the objectness scores, top-ranked region proposals include the most reasonable captions for the objects of the scene. With the comparison of the interactive segmentation result and the bounding boxes, the best match bounding box and the corresponding caption will be obtained (Fig. \ref{fig3}).     
\section{Experiments}
\indent
\textbf{Datasets.}  
For fine tuning of the LFCN, we used the PASCAL VOC 2012 segmentation dataset \cite{D10}. The dataset includes 1464 images for training and 1449 images for validation that are distributed in 20 different classes. We used the whole bunch of these samples to generate our special training pairs in the preprocessing step. For the final validation of the model as well as its comparison with state-of-the-art interactive segmentation, we utilized different well-known segmentation benchmarks including Alpha Matting \cite{AlphaMatting}, Berkeley segmentation dataset (BSDS500) \cite{Berkeley}, Weizmann segmentation evaluation database \cite{Weizmann}, image object segmentation visual quality evaluation database \cite{Inter_[9]} and VOC validation subset.  
\\
\textbf{Preprocessing.}
To generate all the necessary training pairs of the interactive segmentation process, we produced positive and negative Voronoi diagrams with respect to each object that is visible in VOC dataset. The positive seeds are selected randomly inside each object while the MCD approach is used to generate three distinct sets of negative seeds with different distances from the intended object. In the last step, each combination of positive/negative Voronoi diagrams, forms a unique training pair. This leads to production of 97,055 interaction patterns. We preserved 7,055 instances for the test and used the rest as the training data.

\subsection{Fine Tuning of the Proposed LFCN Architecture}
To reach the best quality for the interactive segmentation, our LFCN is trained in three different levels of granularity as proposed in \cite{I17}: LFCN32s, LFCN16s and LFCN8s. 
RGB channels of the input image should be concatenated with the corresponding Voronoi diagrams to form a training instance. Consequently, the first convolutional layer of our LFCN contains five channels. During the network initialization, the RGB-related channels will be initialized by the parameters of the original FCN \cite{I17}. For two extra channels that are associated with Voronoi diagrams, the zero initialization is the best choice as also mentioned in \cite {iter}. Learning parameters of the finer networks should be initialized from the coarser one. The global learning rates of the networks are 1e-8, 1e-10 and 1e-12, respectively while the extended convolutional layers exploit one hundred times bigger learning rates. The learning policy is fixed and we used the weight decay of 5e-3.

\subsection{Metrics}
In order to evaluate UIR localization accuracy of the proposed model, we calculated the well-known measure of Intersection over Union (IoU). To this aim, we computed the IoU of the detected UIR and the corresponding binary label of the validation samples. For the sake of complete comparison between our model and other interactive segmentation techniques, three performance metrics of pixel accuracy, mean accuracy and mean IoU are computed.
\\
The segmentation task of the proposed approach can be considered as a binary segmentation where the classes are limited to foreground (UIR) and background. So, we used binary interpretation of the semantic segmentation metrics that are proposed by Long et al. \cite{I17}:
\begin{itemize}
\item \textbf{Pixel Accuracy (Pixel Acc.):}
\\This measure represents the proportion of the correctly classified foreground ($C_f$) and background ($C_b$) pixels (true positive rates) to the total number of ground truth pixels in foreground ($F$) and background ($B$).
\begin{equation}
\frac{C_f+C_b}{F+B}
\end{equation}
Unfortunately, this metric can be easily influenced by the class imbalance. Hence, high pixel accuracy does not necessarily mean that the accuracy is acceptable when one of the classes is too small or too large.

\item \textbf{Mean Pixel Accuracy (Mean Acc.):}
\\This measure is computed as the mean of the separate foreground and background pixel accuracies:
\begin{equation}
\frac{\frac{C_f}{F}+\frac{C_b}{B}}{{2}}
\end{equation}
This metric alleviates the imbalance problem but can be still misleading. For example when the great majority of pixels are background, a method that predicts all the pixels as background can still have seemingly good performance.
\item \textbf{Mean Intersection over Union (Mean IoU):}
\\Intersection over union is the matching ratio between the result of object localization process and the corresponding ground truth label. This metric is the average of the computed intersection over union for the foreground and background regions:
\begin{equation}
\frac{{(\frac{C_f}{C_f+FP+FN})}_f+{(\frac{C_b}{C_b+FP+FN})}_b}{2}
\end{equation}
Here $FP$ and $FN$ denote the number of the false positive and false negative rates of each class, respectively. This metric solves the previously described issues.
\end{itemize}    
\subsection{Results}
\par
\begin{figure}[!h]
\begin{center}
\begin{tabular}{cccccc}
\hspace{-2mm}\includegraphics[width=0.15\linewidth]{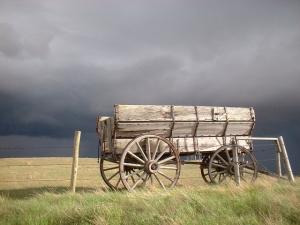}
&
\hspace{-2mm}\includegraphics[width=0.15\linewidth]{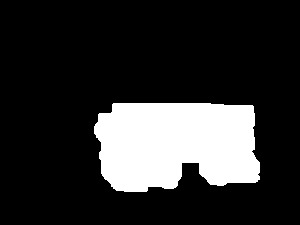}
&
\hspace{-2mm}\includegraphics[width=0.15\linewidth]{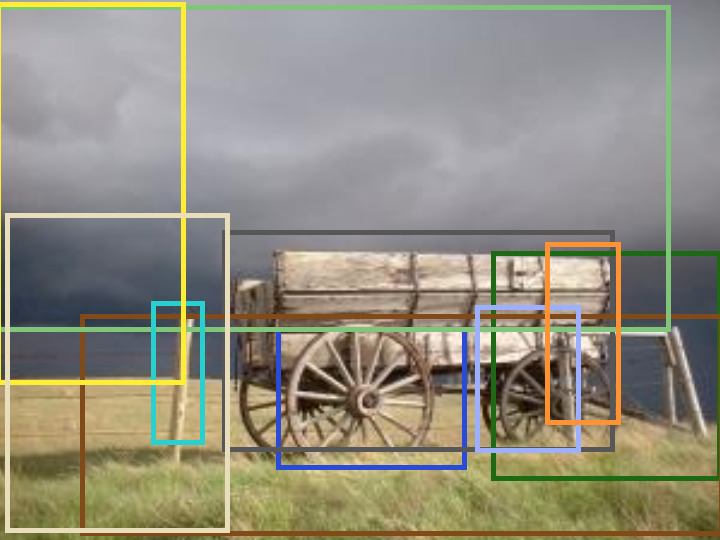}
&
\hspace{-2mm}\includegraphics[width=0.15\linewidth]{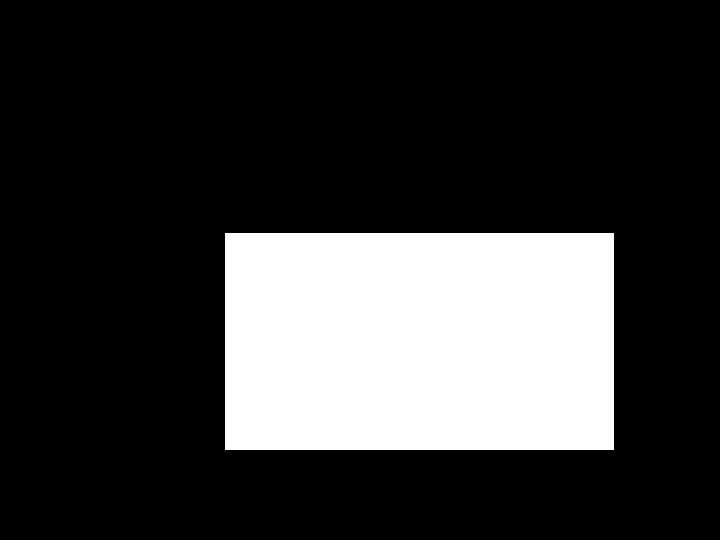}
&
\hspace{-2mm}\includegraphics[width=0.15\linewidth]{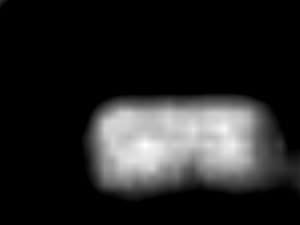}
&
\hspace{-2mm}\includegraphics[width=0.15\linewidth]{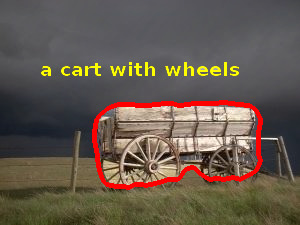}
\\
\hspace{-2mm}\includegraphics[width=0.15\linewidth]{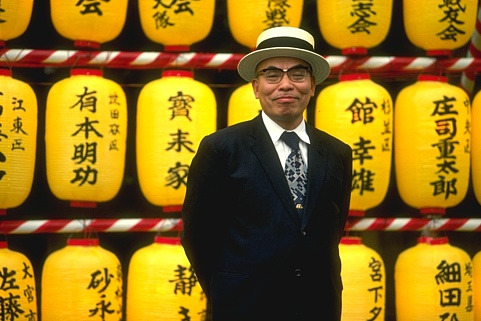}
&
\hspace{-2mm}\includegraphics[width=0.15\linewidth]{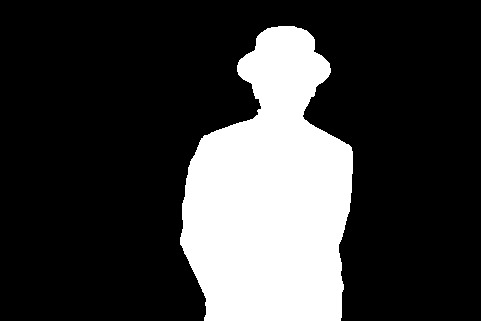}
&
\hspace{-2mm}\includegraphics[width=0.15\linewidth]{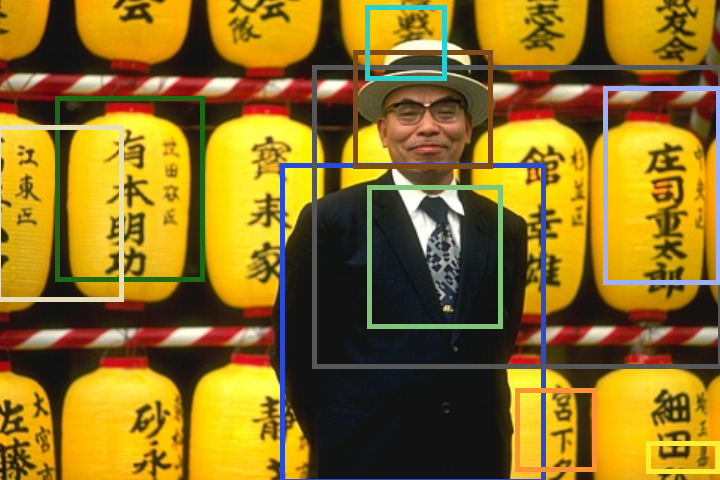}
&
\hspace{-2mm}\includegraphics[width=0.15\linewidth]{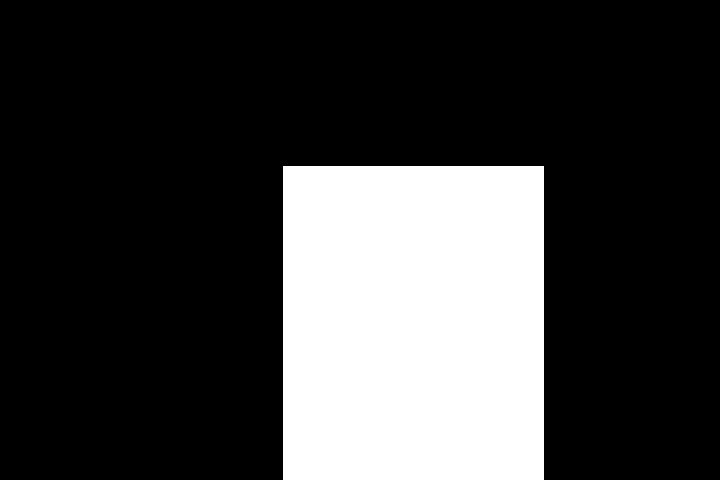}
&
\hspace{-2mm}\includegraphics[width=0.15\linewidth]{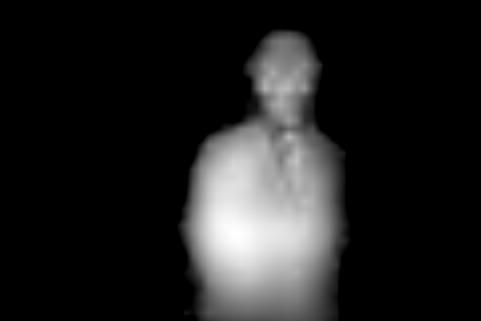}
&
\hspace{-2mm}\includegraphics[width=0.15\linewidth]{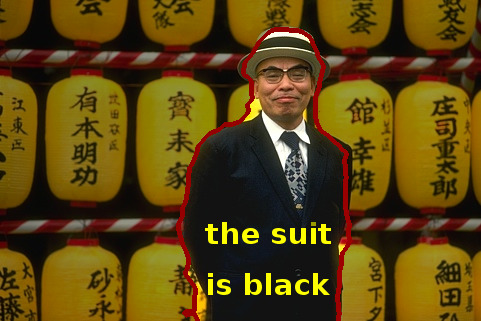}
\\
\hspace{-2mm}\includegraphics[width=0.15\linewidth]{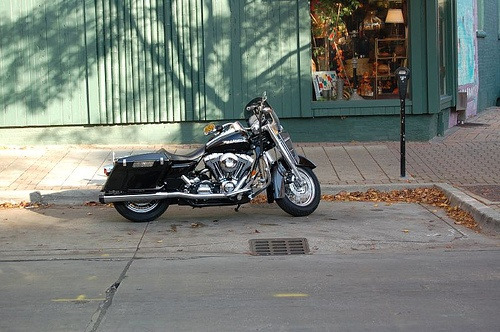}
&
\hspace{-2mm}\includegraphics[width=0.15\linewidth]{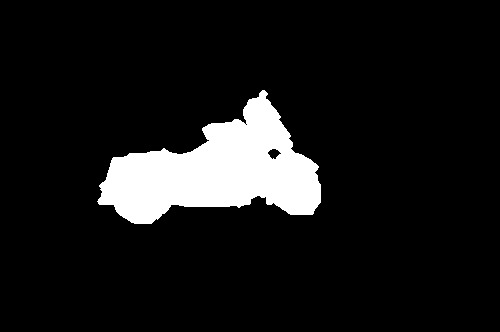}
&
\hspace{-2mm}\includegraphics[width=0.15\linewidth]{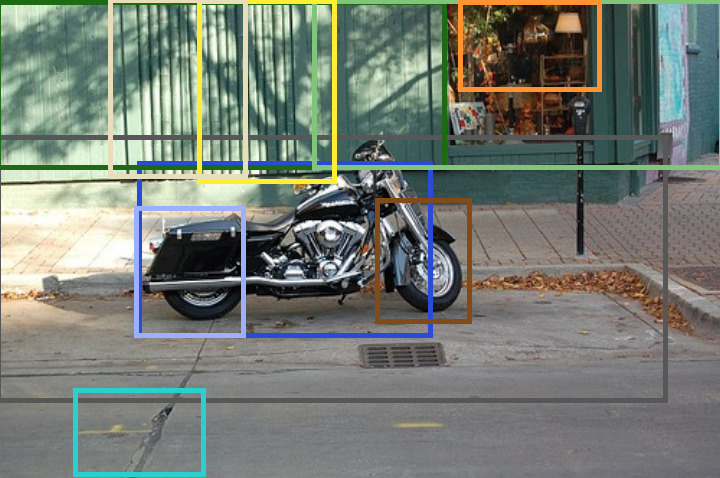}
&
\hspace{-2mm}\includegraphics[width=0.15\linewidth]{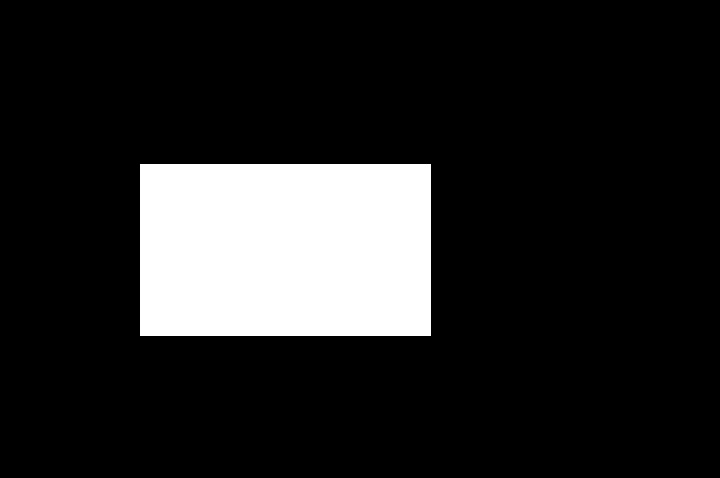}
&
\hspace{-2mm}\includegraphics[width=0.15\linewidth]{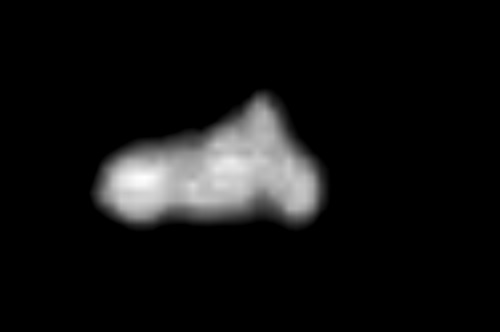}
&
\hspace{-2mm}\includegraphics[width=0.15\linewidth]{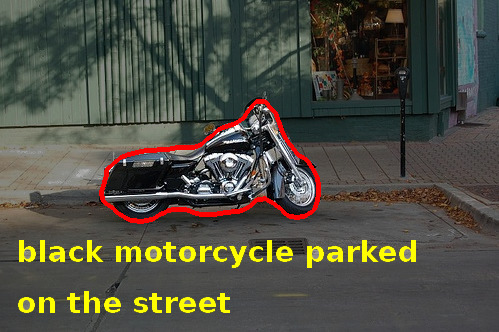}
\\
\hspace{-2mm}\includegraphics[width=0.15\linewidth]{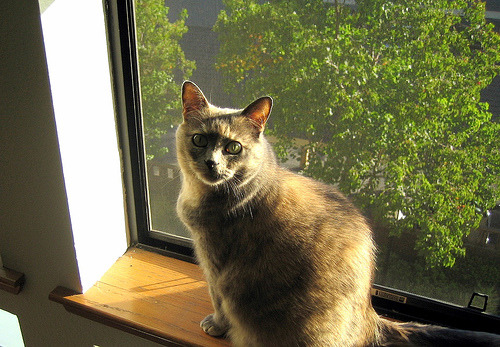}
&
\hspace{-2mm}\includegraphics[width=0.15\linewidth]{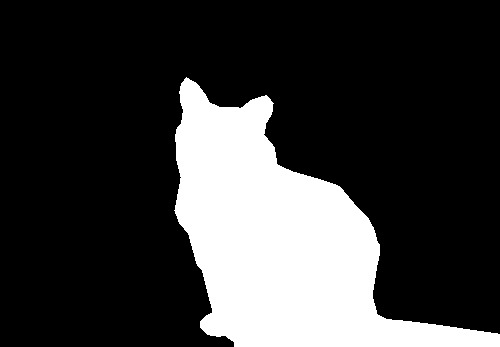}
&
\hspace{-2mm}\includegraphics[width=0.15\linewidth]{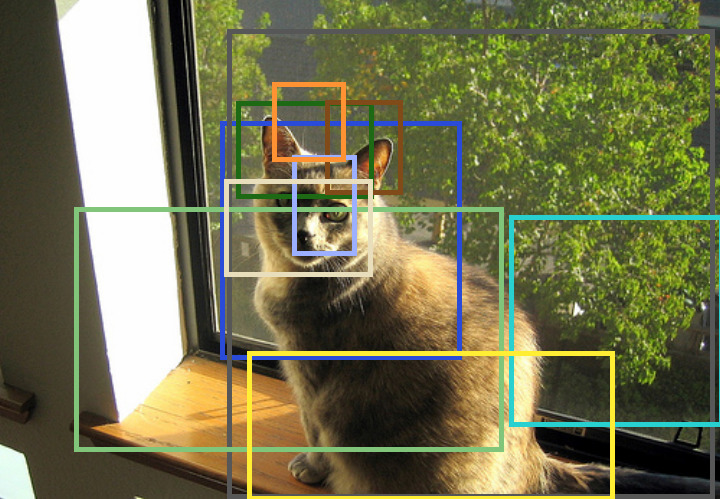}
&
\hspace{-2mm}\includegraphics[width=0.15\linewidth]{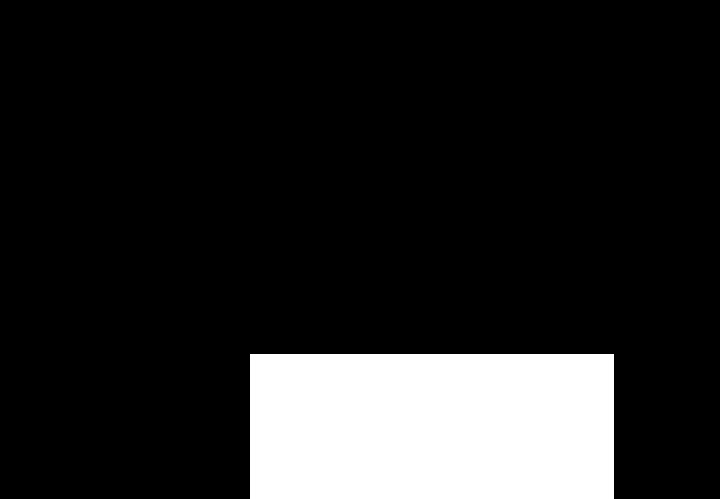}
&
\hspace{-2mm}\includegraphics[width=0.15\linewidth]{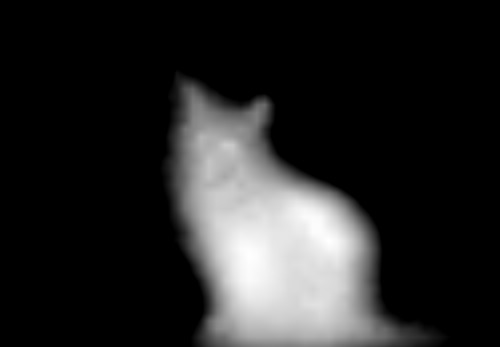}
&
\hspace{-2mm}\includegraphics[width=0.15\linewidth]{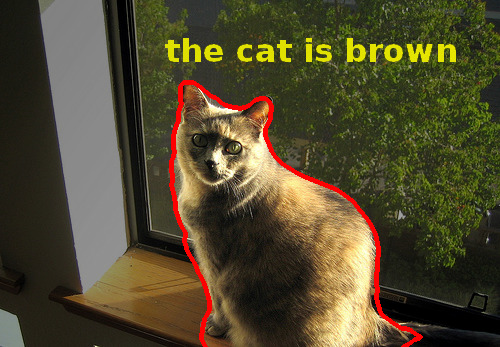}
\\
\hspace{-2mm}\includegraphics[width=0.15\linewidth]{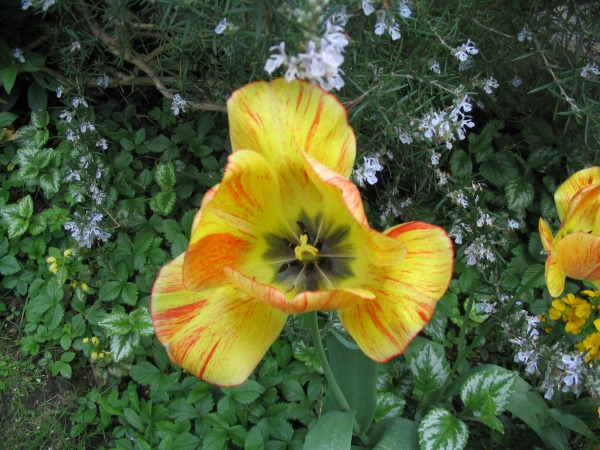}
&
\hspace{-2mm}\includegraphics[width=0.15\linewidth]{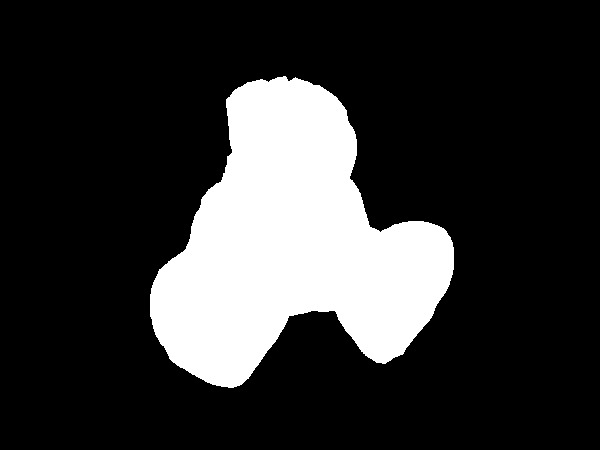}
&
\hspace{-2mm}\includegraphics[width=0.15\linewidth]{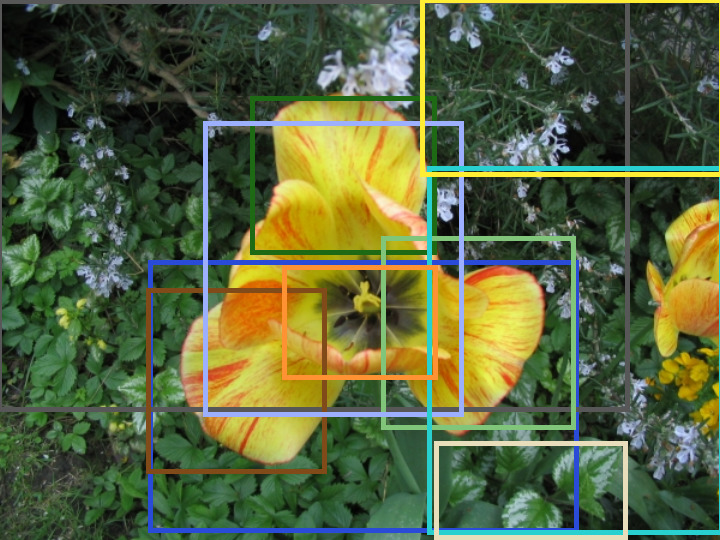}
&
\hspace{-2mm}\includegraphics[width=0.15\linewidth]{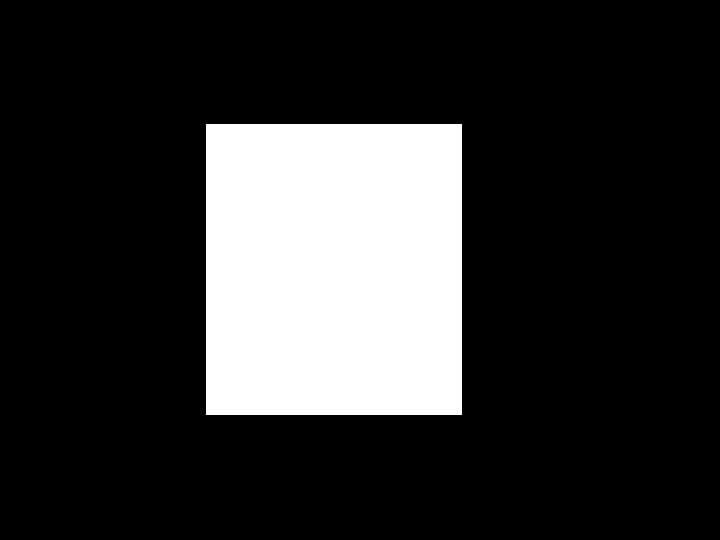}
&
\hspace{-2mm}\includegraphics[width=0.15\linewidth]{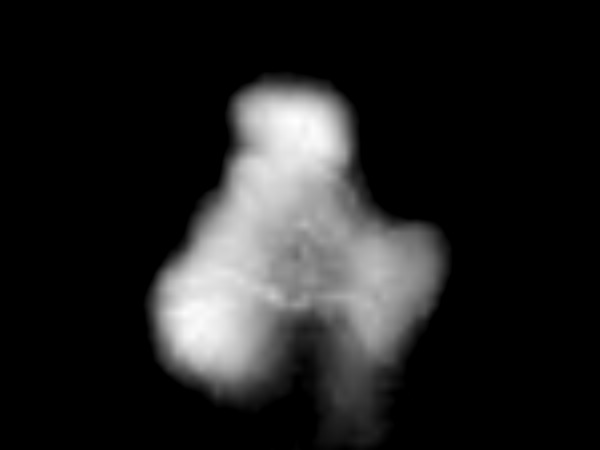}
&
\hspace{-2mm}\includegraphics[width=0.15\linewidth]{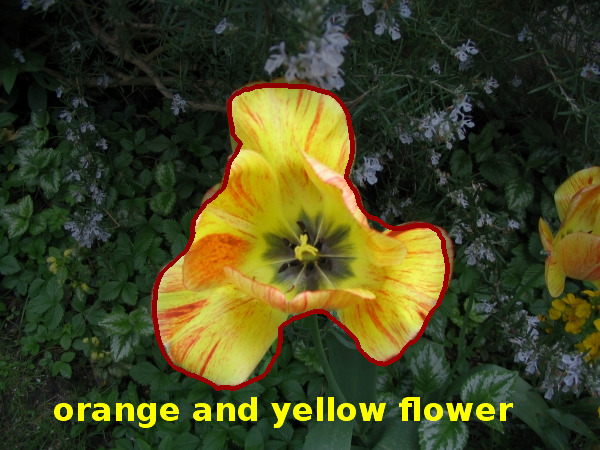}
\\
\end{tabular}
\end{center}
   \caption{\small{From left to right: input images, UIR ground truths, dense captioning bounding boxes \cite{dense}, best match bounding boxes, our LFCN probability maps and the final outputs of our model  including highlighted UIR and its description.}}
\label{fig4}

\end{figure}
\begin{figure}[!h]
\begin{center}
\begin{tabular}{c}
\includegraphics[width=0.6\linewidth]{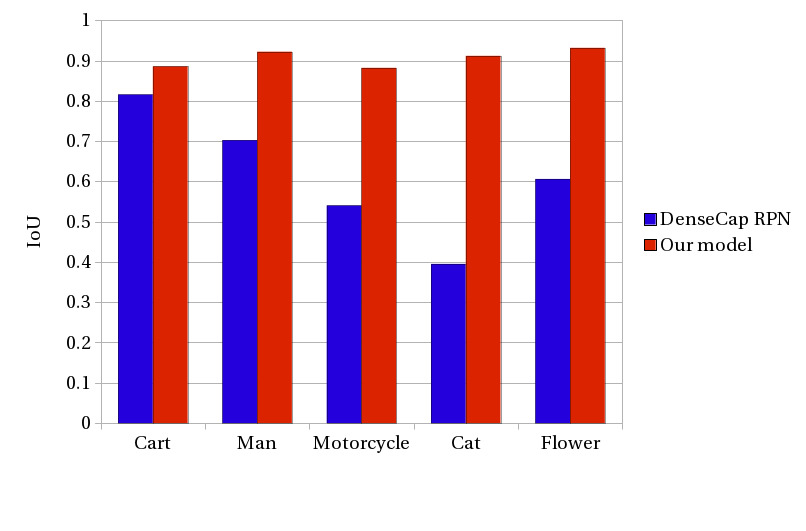}
\\
\includegraphics[width=0.62\linewidth]{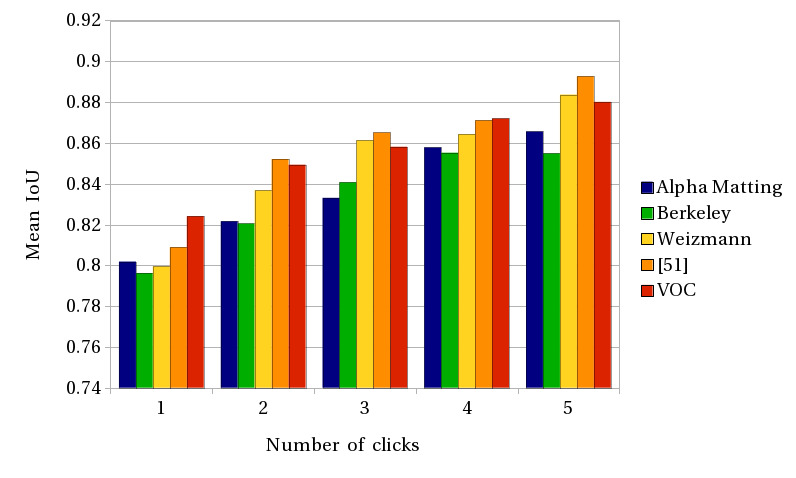}
\\
\end{tabular}
\end{center}
\caption{\small{Localization accuracy comparison between our model and the internal RPN of the DenseCap \cite{dense} (up), and mean IoU accuracy of the proposed model on several segmentation benchmarks against different number of clicks (down).}}
\label{fig5}
\vspace{-5mm}
\end{figure}
\begin{figure}[!h]
\begin{tabular}{l}
\hspace{5mm}
\includegraphics[width=0.6\linewidth]{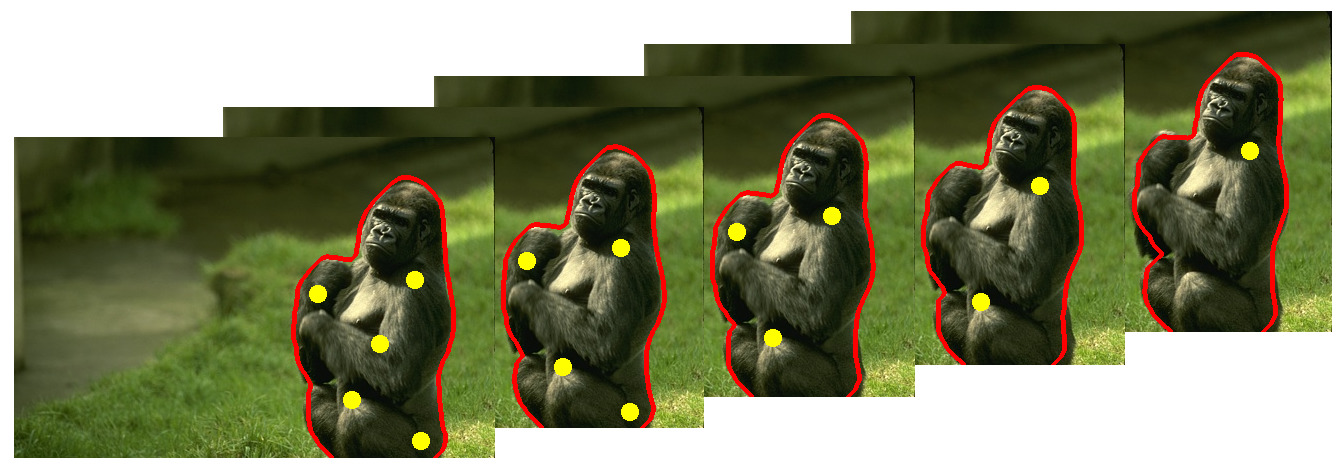}
\end{tabular}
\footnotesize{
\begin{tabular}{c|c}
Number of clicks & IoU \\
\hline
One   & 0.8895 \\
Two   & 0.9141 \\
Three & 0.9227 \\
Four  & 0.9406 \\
Five  & 0.9485
\end{tabular}}
\vspace{1mm}
\caption{\small{Model sensitivity analysis against number of clicks on a sample input image (left), and its associated IoU rates (right).}}
\label{fig6}
\vspace{-2mm}
\end{figure}

\textbf{Test of localization accuracy.} 
In the first step of evaluation, we test our model with a random subset of unseen samples in validation datasets. The response of the model to some instances is shown in Fig. \ref{fig4}. It can be noticed that the output of our approach achieves a considerable rate of accuracy regarding the similarity of the model output with the corresponding ground truth. Furthermore, the confusing output of the dense image captioning is replaced with an explicit situation where the segmented UIR and its description are easily distinguishable.
Fig. \ref{fig5} (left diagram) presents a comparison between the localization accuracy of the proposed method and the internal RPN of the DenseCap \cite{dense} in terms of the obtained IoU for the samples presented in Fig. \ref{fig4}. As illustrated, our model provides a significant improvement regarding the localization accuracy. These results also demonstrate the proficiency of our model in combining interactive segmentation, region proposal and image captioning techniques.
\\
\indent
\textbf{Sensitivity analysis.} 
In this part, we analysed variations of the model output quality against the number of user interactions. As it is shown in Fig \ref{fig6}, although IoU can be improved by applying more user interactions that facilitate boundary detection, our model still provides very good results even by minimum number of clicks. We also provided mean IoU accuracy of the proposed model for five different datasets in Fig. \ref{fig5} (right diagram) that confirms satisfying performance of our model in the case of low interactive information. This noticeable property of our approach makes it convenient to be applied in real-world applications. During our experiments, the proposed method clearly achieves a satisfying segmentation outcome with just one click.
\par
\begin{figure}[!h]
\begin{center}
\begin{tabular}{c c c c c}
\footnotesize{LFCN8s}
& \hspace{-4mm}
\footnotesize{FCN8s}
& \hspace{-4mm}
\footnotesize{GM}
& \hspace{-4mm}
\footnotesize{GrowCut}
& \hspace{-4mm}
\footnotesize{GrabCut}
\\
\includegraphics[width=0.18\linewidth]{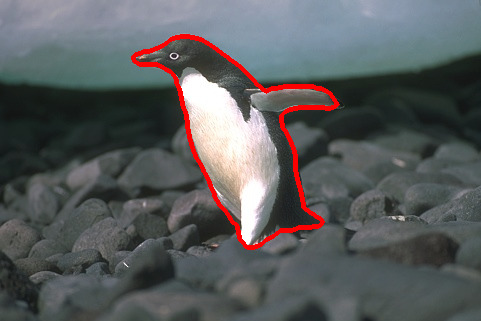}
& \hspace{-4mm}
\includegraphics[width=0.18\linewidth]{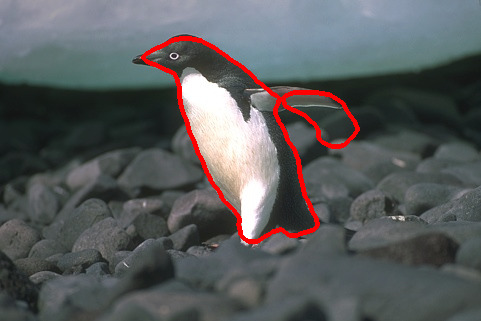}
& \hspace{-4mm}
\includegraphics[width=0.18\linewidth]{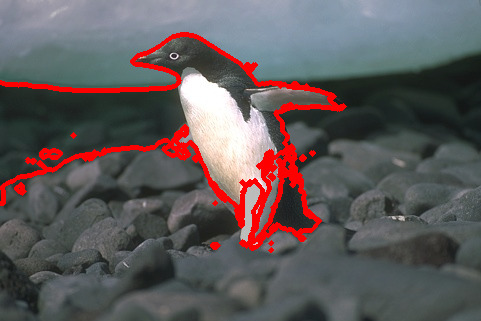}
& \hspace{-4mm}
\includegraphics[width=0.18\linewidth]{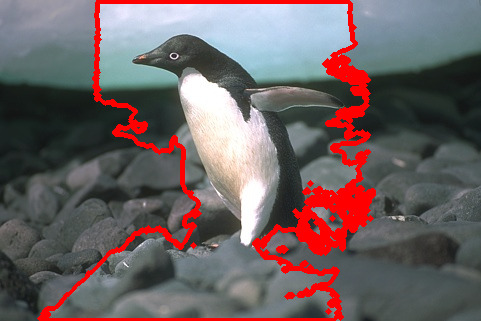}
& \hspace{-4mm}
\includegraphics[width=0.18\linewidth]{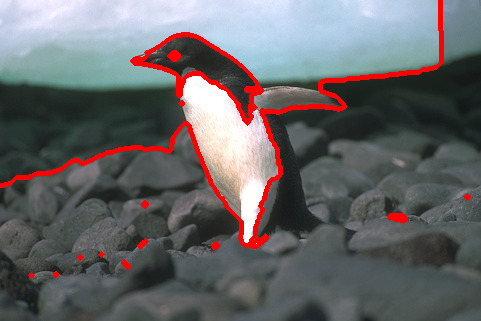}
\\
\includegraphics[width=0.18\linewidth]{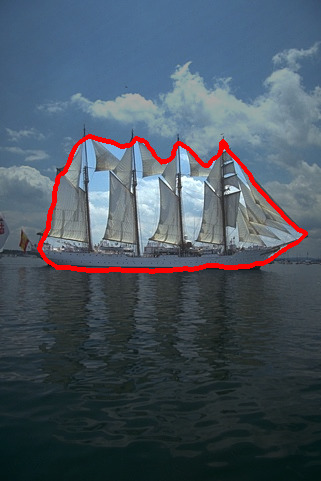}
& \hspace{-4mm}
\includegraphics[width=0.18\linewidth]{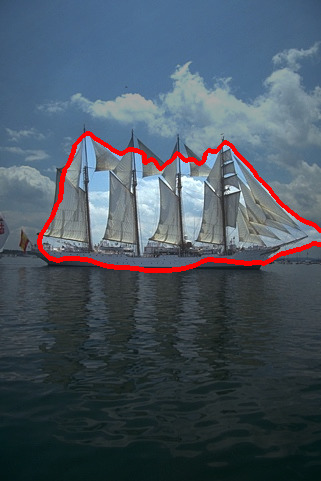}
& \hspace{-4mm}
\includegraphics[width=0.18\linewidth]{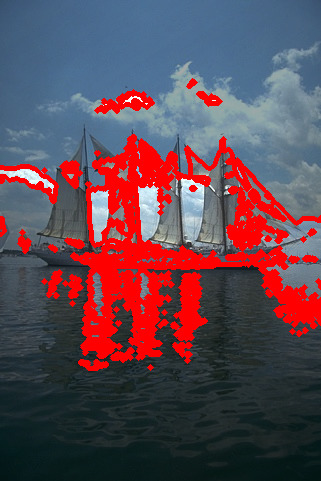}
& \hspace{-4mm}
\includegraphics[width=0.18\linewidth]{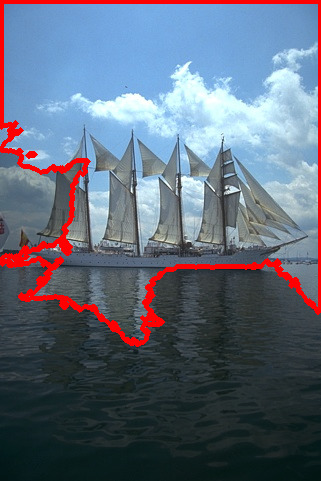}
& \hspace{-4mm}
\includegraphics[width=0.18\linewidth]{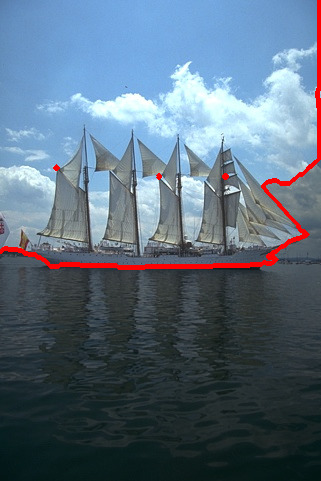}
\end{tabular}
\\
\begin{tabular}{c c c c c c}
\footnotesize{BJ}
& \hspace{-4mm}
\footnotesize{GSC}
& \hspace{-4mm}
\footnotesize{GSCSEQ}
& \hspace{-4mm}
\footnotesize{RW}
& \hspace{-4mm}
\footnotesize{SP}
& \hspace{-4mm}
\footnotesize{MARG}
\\
\includegraphics[width=0.15\linewidth]{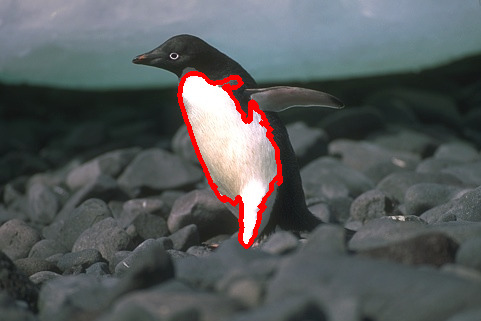}
& \hspace{-4mm}
\includegraphics[width=0.15\linewidth]{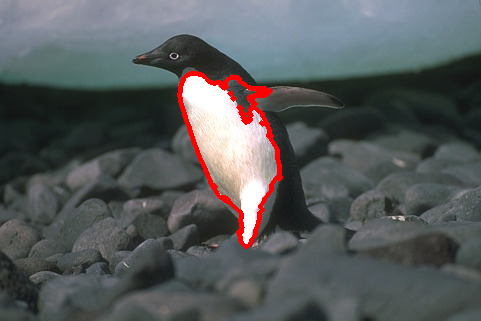}
& \hspace{-4mm}
\includegraphics[width=0.15\linewidth]{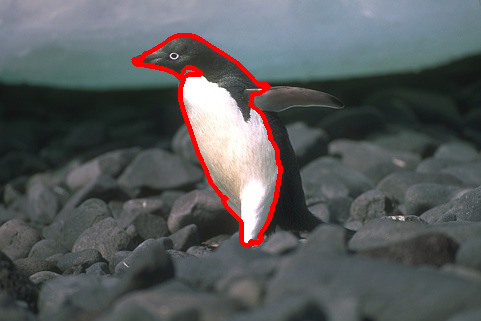}
& \hspace{-4mm}
\includegraphics[width=0.15\linewidth]{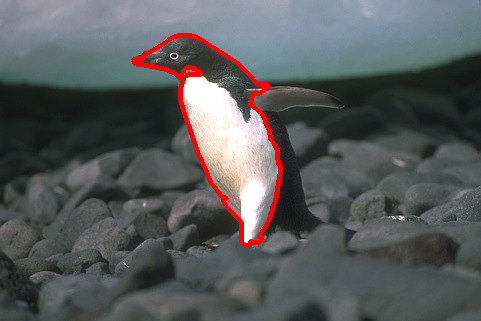}
& \hspace{-4mm}
\includegraphics[width=0.15\linewidth]{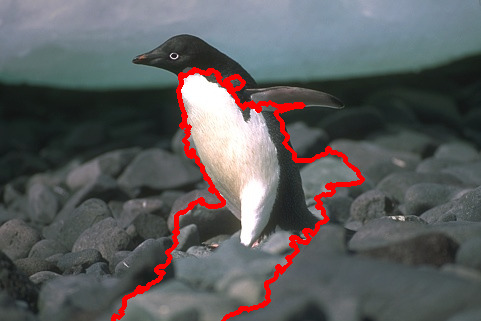}
& \hspace{-4mm}
\includegraphics[width=0.15\linewidth]{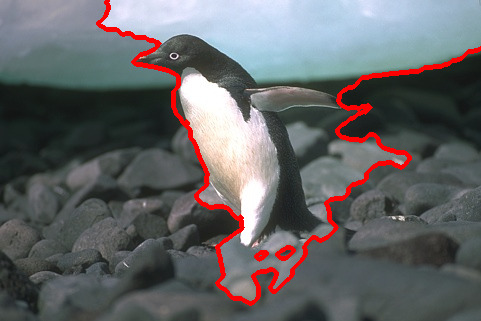}
\\
\includegraphics[width=0.15\linewidth]{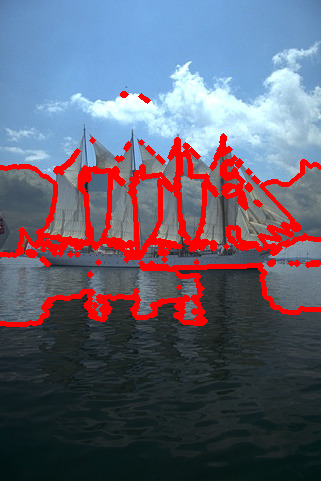}
& \hspace{-4mm}
\includegraphics[width=0.15\linewidth]{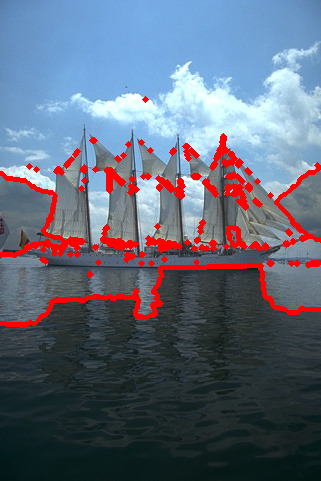}
& \hspace{-4mm}
\includegraphics[width=0.15\linewidth]{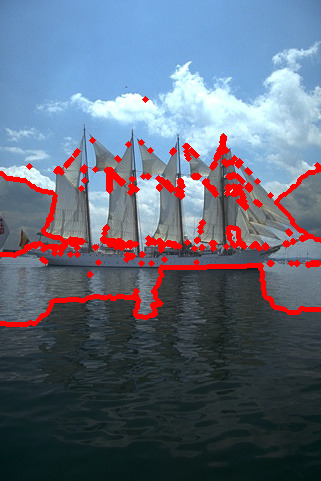}
& \hspace{-4mm}
\includegraphics[width=0.15\linewidth]{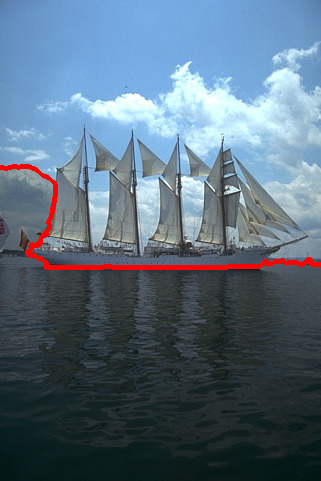}
& \hspace{-4mm}
\includegraphics[width=0.15\linewidth]{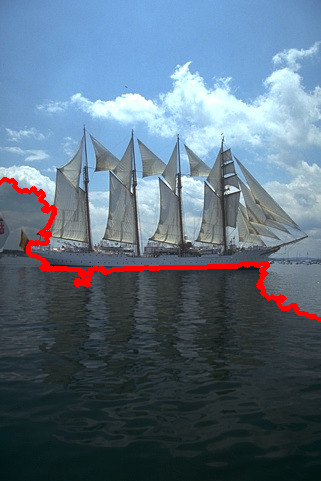}
& \hspace{-4mm}
\includegraphics[width=0.15\linewidth]{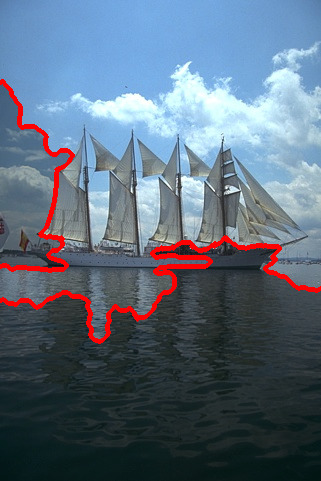}
\end{tabular}
\\
\end{center}
   \caption{\small{Segmentation quality comparison between our model (LFCN8s) and other interactive segmentation techniques. Our method clearly provides higher level of region understanding.}}
\label{fig7}
\vspace{-3mm}
\end{figure}
\begin{figure}[!t]
\begin{center}
\vspace{5mm}
\begin{tabular}{c c c}
\includegraphics[width=0.25\linewidth]{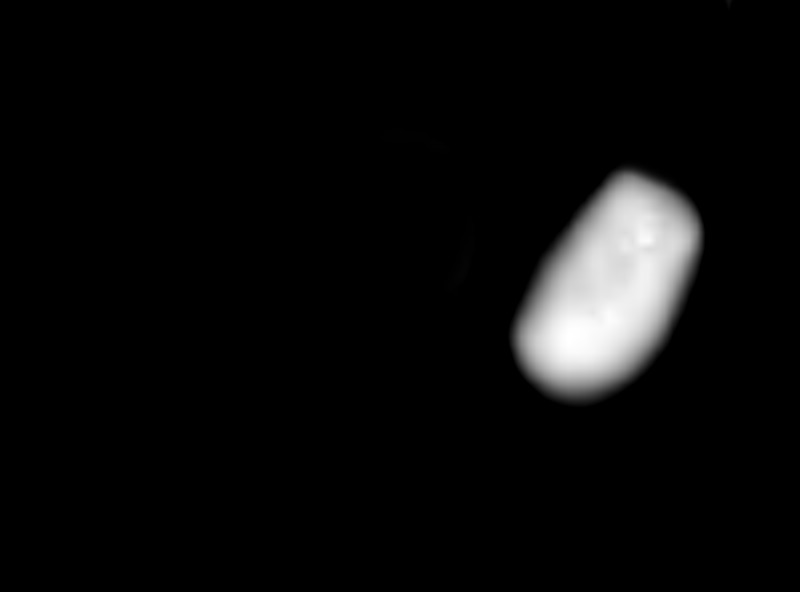}
&
\includegraphics[width=0.25\linewidth]{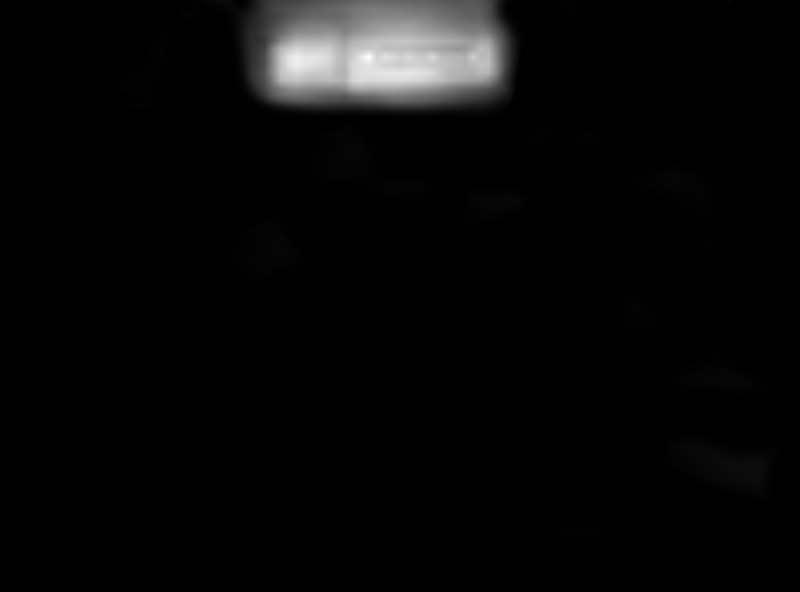}
&
\includegraphics[width=0.25\linewidth]{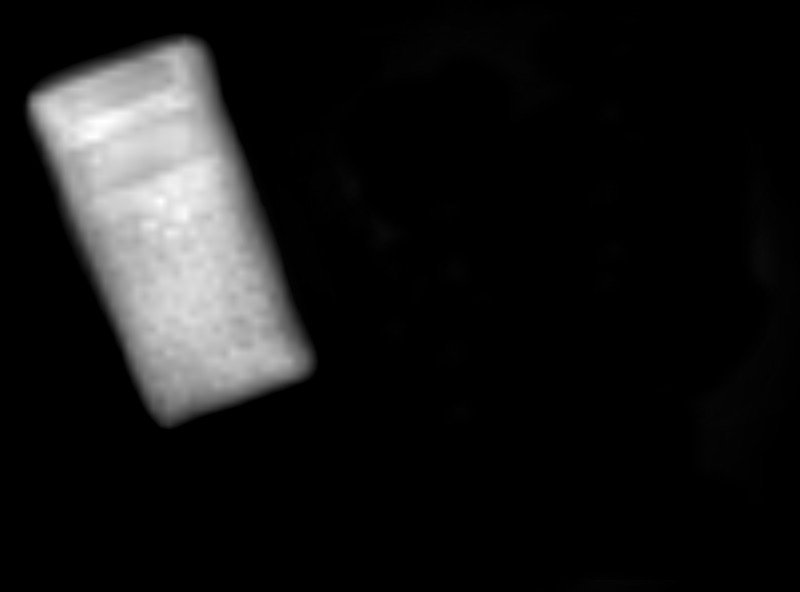}
\\
\includegraphics[width=0.25\linewidth]{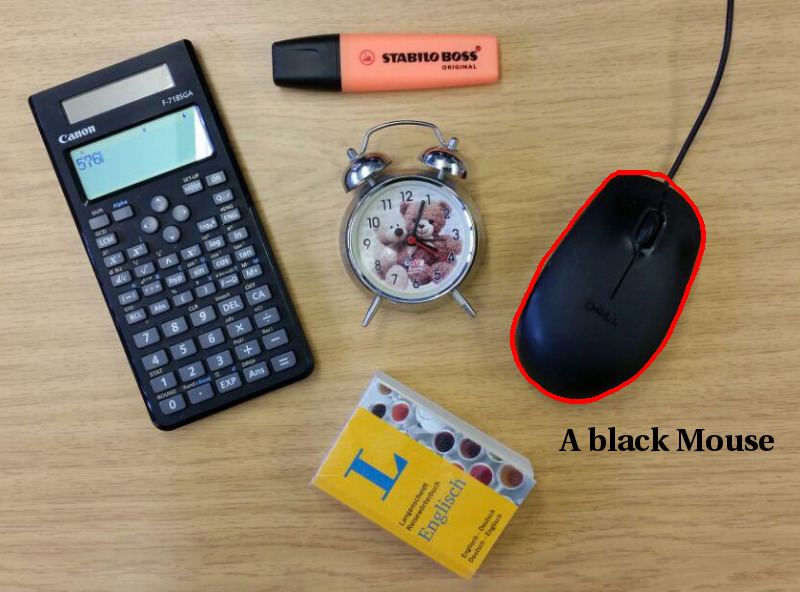}
&
\includegraphics[width=0.25\linewidth]{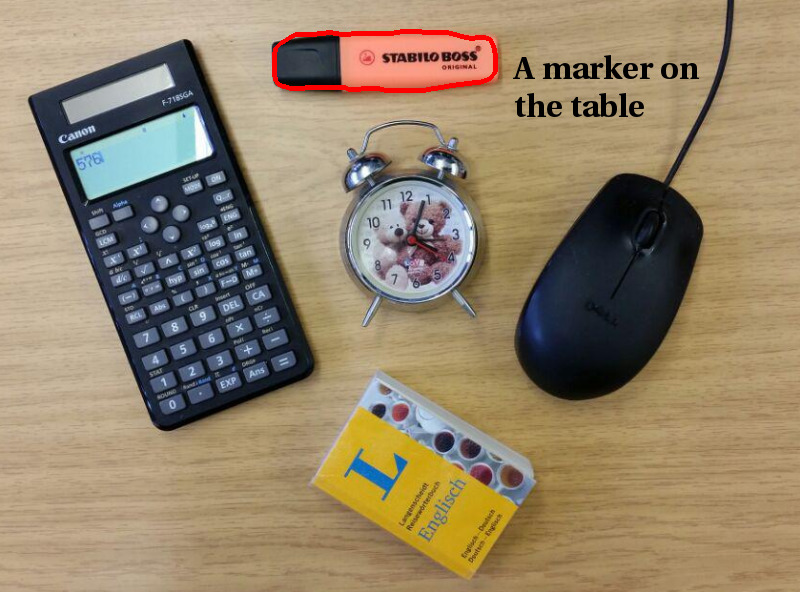}
&
\includegraphics[width=0.25\linewidth]{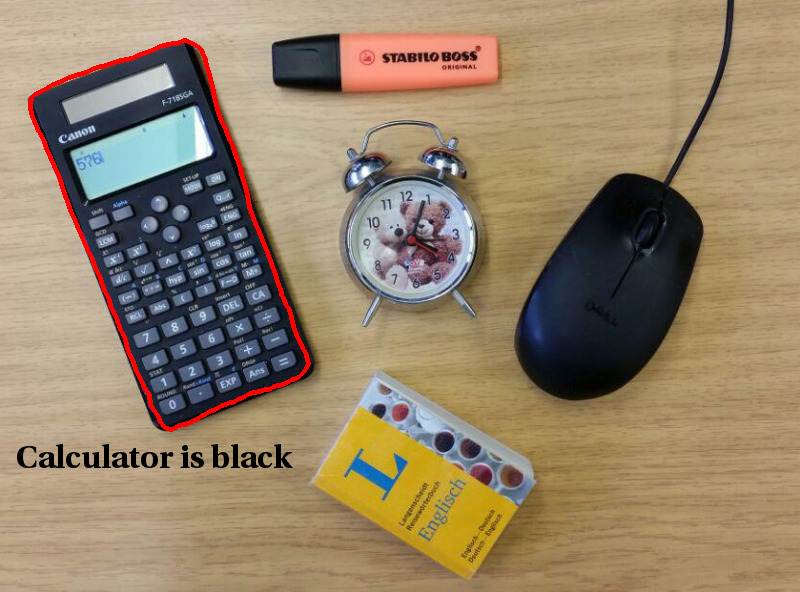}
\\
\end{tabular}
\begin{tabular}{c c}
\includegraphics[width=0.25\linewidth]{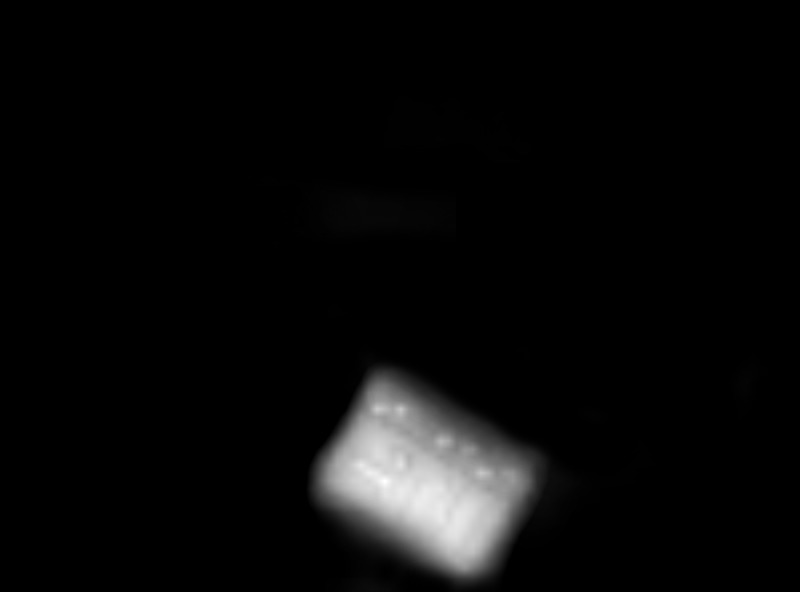}
&
\includegraphics[width=0.25\linewidth]{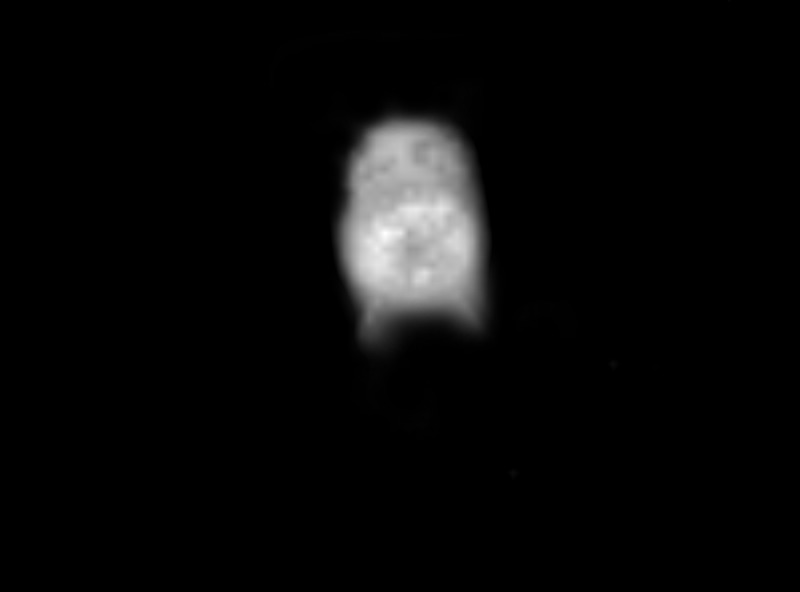}
\\
\includegraphics[width=0.25\linewidth]{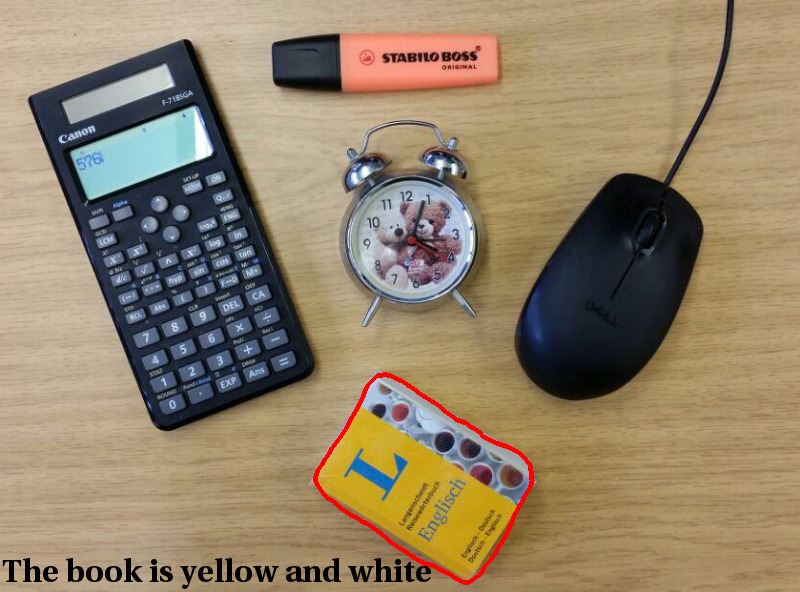}
&
\includegraphics[width=0.25\linewidth]{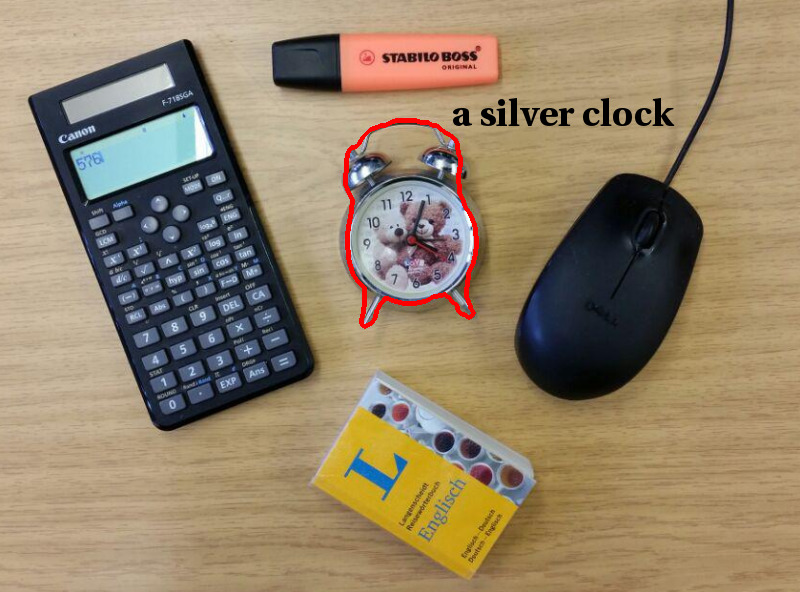}
\\
\end{tabular}
\end{center}
\caption{\small{Probability maps and final outputs of the proposed model in response of user interactions over different parts of an image.}}
\label{fig8}
\end{figure}

\textbf{Comparison of the segmentation quality.} 
In this part we performed an extensive evaluation on segmentation capabilities of the proposed method versus some prevalent segmentation techniques such as Geodesic Matting (GM) \cite{I1}, GrowCut \cite{I26}, Grabcut \cite{GrabCut}, Boykov Jolly (BJ) interactive graph cuts \cite{I2}, Geodesic Star Convexity (GSC) \cite{GSC}, Geodesic Star Convexity with sequential constraints (GSCSEQ), Random Walker (RW) segmentation \cite{I8}, Shortest Path-based interactive segmentation (SP) \cite{SP} and Matching Attributed Relational Graphs (MARG) \cite{MARG}. In all the experiments we generated five positive and five negative clicks randomly. For some of the approaches where the user interactions were defined as points or scribbles, we determined click positions with five-pixel-wide circles. To observe the impact of the extended part of the LFCN on the output quality, we also report all the accuracy measures for the normal version of the FCN as well. Table \ref{tab1} and \ref{tab2} present quantitative results that confirm our approach superiority over several other segmentation techniques on five different benchmarks. As a qualitative comparison, Fig. \ref{fig7} represents final segmentation output of the methods in Table \ref{tab1} for two different samples. As it can be seen, our approach provides the most accurate segmentation result with respect to semantic interpretation of the scene using same number of interactions.
\begin{table}[!h]
\begin{center}
\footnotesize{
\begin{tabular}{l c c c |c c c }
& \multicolumn{3}{c}{Berkeley dataset \cite{Berkeley}} & \multicolumn{3}{c}{Weizmann dataset \cite{Weizmann}}\\
\hline
& Pixel Acc. & Mean Acc. & Mean IoU & Pixel Acc. & Mean Acc. & Mean IoU \\
\hline
GM \cite{I1}          & 0.8237 & 0.7645 & 0.6098 & 0.8328 & 0.7906 & 0.6615 \\
GrowCut \cite{I26}    & 0.6639 & 0.7603 & 0.4520 & 0.6847 & 0.7025 & 0.4795 \\
GrabCut \cite{GrabCut}& 0.6614 & 0.7582 & 0.4511 & 0.6896 & 0.7103 & 0.4818 \\
BJ \cite{I2}          & 0.8138 & 0.8292 & 0.6488 & 0.8196 & 0.8416 & 0.6993 \\
GSC \cite{GSC}        & 0.8240 & 0.8465 & 0.6624 & 0.8253 & 0.8463 & 0.6977 \\
GSCSEQ                & 0.8275 & 0.8425 & 0.6654 & 0.8290 & 0.8482 & 0.7014 \\
RW \cite{I8}          & 0.8691 & 0.8046 & 0.6917 & 0.7953 & 0.7440 & 0.6207 \\
SP \cite{SP}          & 0.7838 & 0.8405 & 0.6121 & 0.7712 & 0.8092 & 0.6272 \\
MARG \cite{MARG}      & 0.8067 & 0.6516 & 0.6180 & 0.7907 & 0.8110 & 0.6354 \\
FCN8s                 & 0.9379 & 0.8766 & 0.8352 & 0.9227 & 0.8978 & 0.8573 \\
LFCN8s                & \textbf{0.9597} & \textbf{0.8989} & \textbf{0.8549} & \textbf{0.9549} & \textbf{0.9316} & \textbf{0.8837} \\ 
\end{tabular}}
\end{center}
\caption{\small{Segmentation accuracy comparison between the proposed method (LFCN8s), different types of interactive segmentation techniques and the original version of the FCN8s \cite{I17} on Berkeley \cite{Berkeley} and Weizmann \cite{Weizmann} segmentation benchmarks.}}
\label{tab1}
\end{table}
\begin{table}[!h]
\begin{center}
\begin{tabular}{ c }
\small{
\begin{tabular}{l c c c }
\multicolumn{4}{c}{Alpha matting dataset \cite{AlphaMatting}} \\
\hline
& Pixel Acc. & Mean Acc. & Mean IoU\\
\hline
GM \cite{I1}          & 0.7958 & 0.7993 & 0.6602 \\
GrowCut \cite{I26}    & 0.7924 & 0.7982 & 0.6465 \\
GrabCut \cite{GrabCut}& 0.7908 & 0.7953 & 0.6513 \\
BJ \cite{I2}          & 0.9117 & 0.9119 & 0.8569 \\
GSC \cite{GSC}        & 0.9076 & 0.9056 & 0.8476 \\
GSCSEQ                & 0.9092 & 0.9066 & 0.8492 \\
RW \cite{I8}          & 0.8568 & 0.8599 & 0.7561 \\
SP \cite{SP}          & 0.8390 & 0.8466 & 0.7258 \\
MARG \cite{MARG}      & 0.9131 & 0.9152 & 0.8434 \\
FCN8s                 & 0.9193 & 0.9068 & 0.8438 \\
LFCN8s                & \textbf{0.9320} & \textbf{0.9227} & \textbf{0.8656} \\ 
\end{tabular}}
\\
\\
\begin{tabular}{l c c c }
\multicolumn{4}{c}{Image object segmentation visual quality evaluation dataset \cite{Inter_[9]}}\\
\hline
& Pixel Acc. & Mean Acc. & Mean IoU\\
\hline
GM \cite{I1}          & 0.8313 & 0.7731 & 0.6226 \\
GrowCut \cite{I26}    & 0.6600 & 0.6968 & 0.4425 \\
GrabCut \cite{GrabCut}& 0.6571 & 0.6922 & 0.4391 \\
BJ \cite{I2}          & 0.8108 & 0.8069 & 0.6406 \\
GSC \cite{GSC}        & 0.8170 & 0.8087 & 0.6458 \\
GSCSEQ                & 0.8193 & 0.8092 & 0.6486 \\
RW \cite{I8}          & 0.7787 & 0.7665 & 0.5810 \\
SP \cite{SP}          & 0.7914 & 0.8030 & 0.6053 \\
MARG \cite{MARG}      & 0.7651 & 0.7871 & 0.5685 \\
FCN8s                 & 0.9526 & 0.9155 & 0.8710 \\
LFCN8s                & \textbf{0.9689} & \textbf{0.9332} & \textbf{0.8927} \\ 
\end{tabular}
\\
\\
\begin{tabular}{l c c c }
\multicolumn{4}{c}{VOC validation dataset \cite{D10}} \\
\hline
& Pixel Acc. & Mean Acc. & Mean IoU\\
\hline
GM \cite{I1}          & 0.8165 & 0.7283 & 0.5787 \\
GrowCut \cite{I26}    & 0.6268 & 0.6366 & 0.3999 \\
GrabCut \cite{GrabCut}& 0.6282 & 0.6412 & 0.4028 \\
BJ \cite{I2}          & 0.7559 & 0.7824 & 0.5794 \\
GSC \cite{GSC}        & 0.7707 & 0.7879 & 0.5903 \\
GSCSEQ                & 0.7724 & 0.7883 & 0.5912 \\
RW \cite{I8}          & 0.7014 & 0.7102 & 0.5095 \\
SP \cite{SP}          & 0.7602 & 0.7809 & 0.5618 \\
MARG \cite{MARG}      & 0.7118 & 0.7218 & 0.5050 \\
FCN8s                 & 0.9527 & 0.9201 & 0.8723 \\
LFCN8s                & \textbf{0.9630} & \textbf{0.9260} & \textbf{0.8801} \\ 
\end{tabular}
\\
\\
\\
\end{tabular}
\end{center}
\caption{\small{Segmentation accuracy comparison between the proposed method (LFCN8s), different types of interactive segmentation techniques and the original version of the FCN8s \cite{I17} on three more segmentation benchmarks.}}
\label{tab2}
\end{table}
\newpage
\noindent
\textbf{Dense interactive region captioning.} In the final part of our experiments, we verified the ability of the proposed model to caption several regions of the image. Fig. \ref{fig8} shows the result of such an experiment where multiple objects in different scales are detected via user interactions and described properly.
\section {Conclusion}
In this paper, we presented a novel hybrid deep learning framework which is capable of targeted segmentation and captioning as a response to user interactive actions. A wide variety of experiments confirmed our model superiority over various state-of-the-art interactive segmentation approaches. In addition, further experiments demonstrated our model capability to caption an arbitrary region of the image with one or few clicks, which is especially convenient for real-world interactive applications. 
\newpage
\bibliography{my_egbib}

\end{document}